\documentclass{article}
\usepackage{ijcai24}

\usepackage{times}
\usepackage{soul}
\usepackage{url}
\usepackage[hidelinks]{hyperref}
\usepackage[utf8]{inputenc}
\usepackage[small]{caption}
\usepackage{graphicx}
\usepackage{amsmath}
\usepackage{amsthm}
\usepackage{booktabs}
\usepackage{algorithm}
\usepackage{algorithmic}
\usepackage[switch]{lineno}

\usepackage{amsmath}
\usepackage{amssymb}
\usepackage{mathtools}
\usepackage{amsthm}

\DeclareMathOperator{\argmax}{arg\,max\,}
\DeclareMathOperator{\argmin}{arg\,min\,}

\usepackage{multirow}

\usepackage{comment}

\usepackage{float}

\usepackage{xcolor}

\title{A Deep Probabilistic Spatiotemporal Framework for Dynamic Graph Representation Learning with Application to Brain Disorder Identification}

\author{
Sin-Yee Yap$^{1,2,*}$
\and
Junn Yong Loo$^{1,*,\dagger}$\and
Chee-Ming Ting$^{1,*}$\and
Fuad Noman$^1$\and
Rapha\"{e}l C.-W. Phan$^1$ \and
Adeel Razi$^3$ \And
David L. Dowe$^2$ \\
\affiliations
$^1$School of Information Technology, Monash University Malaysia\\
$^2$Department of Data Science and AI, Faculty of Information Technology, Monash University, Australia\\
$^3$Turner Institute for Brain and Mental Health, School of Psychological Sciences, Monash University\\
\thanks{These authors contributed equally to this work.}
\thanks{Corresponding author.}
\emails
\{sin.yap, loo.junnyong, ting.cheeming, fuad.noman, raphael.phan, adeel.razi, david.dowe\}@monash.edu
}

\begin{document}


\maketitle

\begin{abstract}
    Recent applications of pattern recognition techniques on brain connectome classification using functional connectivity (FC) are shifting towards acknowledging the non-Euclidean topology and dynamic aspects of brain connectivity across time. In this paper, a deep spatiotemporal variational Bayes (DSVB) framework is proposed to learn time-varying topological structures in dynamic FC networks for identifying autism spectrum disorder (ASD) in human participants. The framework incorporates a spatial-aware recurrent neural network with an attention-based message passing scheme to capture rich spatiotemporal patterns across dynamic FC networks. To overcome model overfitting on limited training datasets, an adversarial training strategy is introduced to learn graph embedding models that generalize well to unseen brain networks. Evaluation on the ABIDE resting-state functional magnetic resonance imaging dataset shows that our proposed framework substantially outperforms state-of-the-art methods in identifying patients with ASD. Dynamic FC analyses with DSVB-learned embeddings reveal apparent group differences between ASD and healthy controls in brain network connectivity patterns and switching dynamics of brain states. The code is available at \url{https://github.com/Monash-NeuroAI/Deep-Spatiotemporal-Variational-Bayes}.
\end{abstract}

\section{Introduction}

\begin{figure*}[!t]
\centerline{\includegraphics[width=\textwidth]{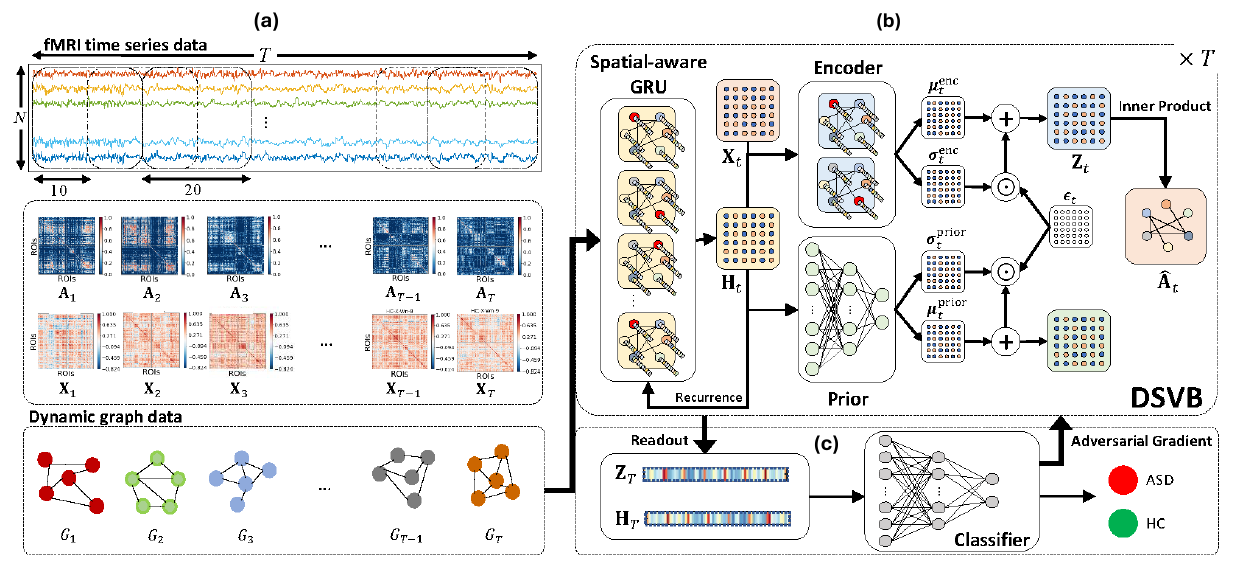}}
\caption{The architecture of the proposed DSVB framework for dynamic functional brain network classification. (a) Dynamic brain network is constructed from sliding window analysis of fMRI time series data. (b) DSVB framework integrating variational Bayes and recurrent layers to learn the sequential stochastic graph latent embeddings. (c) The alterations in latent embeddings readout are used to identify patients with Autism Spectrum Disorder (ASD) through a fully-connected classifier.}
\label{fig:dgae}
\end{figure*}

The human brain is a complex system that consists of numerous interconnected neuronal regions. As revealed by functional magnetic resonance imaging (fMRI), spontaneous spatiotemporal fluctuations in brain activity exist even during a resting state. Functional connectivity (FC) in brain networks is typically characterized via statistical dependence (such as correlations) between blood oxygen level dependent (BOLD) fMRI signals over spatially-distinct brain regions. Alterations in resting-state FC networks have been associated with neuropsychiatric or neurodevelopmental disorders \cite{Filippi2019-rg}, for example in autism spectrum disorder (ASD) \cite{Holiga2019-rf,Valenti2020-yg,wang2021resting}.

Machine learning (ML) techniques have been applied to the identification and prediction of neuropsychiatric disorders in the past decade \cite{Tanveer2020-rf,Hyde2019-vk}. Traditional ML methods such as support vector machines (SVM), artificial neural networks, and deep learning methods are very widely used ML techniques \cite{Moridian2022-lf}. Deep learning has especially gained popularity in the studies of neuroscience as it offers an opportunity to discover and understand the underlying differences in the pattern of brain connectivity between individuals with psychiatric disorders and healthy controls using deep neural networks \cite{RaviPrakash2019-ib,Khodatars2021-ip,Zhang2020-ad}. Conventional approaches have modelled brain networks as 2D grid-like connectivity matrices, neglecting the non-Euclidean nature of brain networks. To address this, graph-based deep learning offers flexibility in modeling individual-level pairwise brain region interactions \cite{Yan2019-az}, capturing group-level associations using population graphs \cite{Jiang2020-mn}, or combining both strategies \cite{Li2022-rr,Zhou2021-uo}. Yet, studies using graph neural networks (GNNs) in these contexts often focus on group-level network topology with phenotype data or supervised subject-level embedding using pre-computed population graphs for classifying static brain networks.

Dynamic spatiotemporal patterns in resting-state brain functional networks have been reported in several studies \cite{Lee2019-tz,Aedo-Jury2020-jb}. They emphasize the consideration of not only the stationary brain network topology but also the dynamic temporal structure. Studies show that GNNs have potential in studying brain networks. However, the application of GNNs to model graph-structured brain network data has predominantly been limited to static representations, and there is a notable scarcity of GNN models designed to investigate the dynamic aspects of such networks.
For instance, \cite{azevedo2020deep} employed 1-dimensional CNNs for temporal feature extraction and graph convolutional networks (GCNs)
for spatial information sharing, and the simple framework showed great potential on both binary and continuous label prediction. \cite{kim2021learning} proposed self-attention and squeeze-excitation for readout, applied successively on GRU-encoded graphs to obtain spatiotemporal embeddings for downstream gender classification. Rather than focusing solely on the gender classification problem, further works \cite{fuad2022,Lingwen2023} incorporated GCNs to initially learn spatial-aware graph embeddings of the dynamic FC. This was followed by a long short-term memory (LSTM) model to leverage the temporal dependencies of these learned graph embeddings for subject-level network classification of psychiatric disorders. These methods have predominantly relied on deterministic models, and their compartmentalized, task-specific modules hinder the retention of full spatiotemporal information across layers.

In this paper, a deep spatiotemporal variational Bayes (DSVB) framework is proposed to learn the stochastic dynamic graph embeddings of FC networks, composed of sliding window segments of the resting-state fMRI time-series data. Figure 1 shows an overview of the proposed DSVB framework.
Derived from general sequential variational Bayes (VB), our proposed framework provides a stochastic representation modelling solution of using dynamic FC networks for neuropsychiatric disorder diagnosis. Unlike existing GNN-based approaches that learn independent and identically distributed (i.i.d.) graph embeddings, we incorporate a spatial-aware recurrent neural network with attention-based message-passing scheme to 
model both the complex topological structure and temporal dependencies within a dynamic FC network sequence.
Additionally, we introduce an adversarial training strategy to overcome the overfitting issue aggravated by limited training data. Our contributions are highlighted as follows:
\begin{enumerate}
\item An end-to-end probabilistic framework, integrating variational Bayes and recurrent GNNs, to learn the sequential stochastic graph latent embedding spaces in a purely unsupervised manner. This accommodates a wider range of spatiotemporal variability across unseen brain FC networks. 
\item A spatial-aware Gated Recurrent Unit (GRU) with attention-based message-passing scheme is incorporated to generate hierarchical latent embeddings to extract underlying non-Euclidean topological structure and rich temporal patterns of dynamic FC networks. 
The alterations in these latent embeddings are then exploited for downstream ASD identification.
\item An adversarial model perturbation strategy is introduced to apply regularization on the GNN-based graph embedding models during training. This adversarial regularization facilitates learning of a smooth latent embedding space that extrapolates well to unseen dynamic FC networks beyond the training dataset.
\item A nested-stratified 5-fold cross-validation shows that our proposed DSVB framework outperforms state-of-the-art methods in ASD identification from fMRI. 
\end{enumerate}

\section{Methods}

\subsection{Dynamic Brain Network Construction}

Dynamic brain graphs were constructed from fMRI data using the sliding-window technique. The regions of interest (ROI) fMRI time series was segmented into multiple overlapping blocks using sliding windows of size $L=20$ with a shift of 10 time steps. To represent the dynamic FC networks, we computed the correlation matrix between ROI time series for each time window. The Ledoit-Wolf (LDW) regularized shrinkage covariance estimator is used to ensure well-conditioned FC estimates.

The dynamic brain network at each time-step $t$ is represented by a graph $G_{t} \equiv (V,E)$, where $v_{i}\in V$ represents a particular brain ROI and $e_{ij} \in E$ is the connectivity edge between a pair of nodes $v_{i}$ and $v_{j}$. The topological structure of dynamic brain networks $G_{t}$ of $N$ (the number of) nodes can be represented by a sequence of time-resolved adjacency matrices $\mathbf{A}_{t}=[a_{t,ij}]\in \{0, 1\}^{N\times N}$
(for each $t$), which were constructed by proportional thresholding of the sliding-windowed correlation-based FC matrices to keep the strongest $40\%$ of connections based on absolute correlation value, and setting other connections to zero. Each dynamic graph $G_{t}$ has associated node features $\mathbf{X}_{t}=[\mathbf{x}_{t,1},\ldots,\mathbf{x}_{t,N}]^{\top} \in \mathbb{R}^{N \times D_x}$ where $\mathbf{x}_{t,i} \in \mathbb{R}^{D_x}$ is the feature (column) vector of node $v_i$. We used the connection weights of each node for $\mathbf{x}_{t,i}$.

\subsection{Spatiotemporal Variational Bayes}

To implicitly learn the spatiotemporal structure of dynamic FC, we consider the unsupervised learning task of reconstructing the sequential network-derived adjacency matrices $\mathcal{A} = \{\mathbf{A}_{t}\}_{t=0}^{T}$ from node feature data $\mathcal{X} = \{\mathbf{X}_{t}\}_{t=0}^{T}$, where $T + 1$ is the sequence length. Under the sequential Bayesian framework, this is achieved via maximizing the variational evidence lower bound (ELBO) of conditional adjacency log-likelihood $\log p_{\theta}(\mathcal{A}|\mathcal{X})$. 

Taking into account the difficulty of directly modelling the complex graph log-likelihood, we introduce a sequence of latent embeddings $\mathcal{Z} = \{\mathbf{Z}_{t}\}_{t=0}^{T}$ with $\mathbf{Z}_{t}=[\mathbf{z}_{t,1},\ldots,\mathbf{z}_{t,N}]^{\top} \in \mathbb{R}^{N\times D_z}$ where $\mathbf{z}_{t,i} \in \mathbb{R}^{D_z}$, and consider an importance decomposition of the ELBO:
\begin{align} \label{eq:ELBO_general}
\begin{split}
\mathcal{L}^\mathrm{ELBO}(\theta,\vartheta) 
&= \mathbb{E}_{q_{\vartheta}(\mathcal{Z}|\mathcal{X},\mathcal{A})}\left[\log \frac{p_{\theta}(\mathcal{A}, \mathcal{Z}|\mathcal{X})}{q_{\vartheta}(\mathcal{Z}|\mathcal{X},\mathcal{A})}\right]
\end{split}
\end{align}
where the subscripts $\theta$ and $\vartheta$ denote GNN parameters that model the generative (prior and decoder) distribution $p_{\theta}(\mathcal{A}, \mathcal{Z}|\mathcal{X})$ and the approximate posterior (encoder) distribution $q_{\vartheta}(\mathcal{Z}|\mathcal{X},\mathcal{A})$
which we can factorize, respectively as 
\begin{align}
\begin{split}
\label{eq:sELBO_factorization}
p_{\theta}(\mathcal{A}, \mathcal{Z}|\mathcal{X}) &= \prod_{t=0}^{T} \;
\overbrace{p_{\theta}(\mathbf{A}_{t}|\mathbf{Z}_{\leq t},\mathbf{X}_{< t},\mathbf{A}_{< t})}^{\text{decoder}} \; \times \\ 
&\qquad\;\;\; \underbrace{p_{\theta}(\mathbf{Z}_{t}|\mathbf{X}_{< t},\mathbf{A}_{< t},\mathbf{Z}_{< t})}_{\text{prior}} \\
q_{\vartheta}(\mathcal{Z}|\mathcal{X},\mathcal{A}) &= \prod_{t=0}^{T} \;
\underbrace{q_{\vartheta}(\mathbf{Z}_{t}|\mathbf{X}_{\leq t},\mathbf{A}_{\leq t},\mathbf{Z}_{< t})}_{\text{encoder}}
\end{split}
\end{align}
Based on the ancestral factorization above, we can then expand (\ref{eq:ELBO_general}) to obtain a sequential ELBO (sELBO) as follows:
\begin{align} \label{eq:sELBO_general}
\begin{split}
& \mathcal{L}^\mathrm{sELBO}(\theta,\vartheta) = \\
& \sum_{t=0}^{T} \Big( \mathbb{E}_{q_{\vartheta}(\mathbf{Z}_{t}|\mathbf{X}_{\leq t},\mathbf{A}_{\leq t},\mathbf{Z}_{< t})}\big[ \log p_{\theta}(\mathbf{A}_{t}|\mathbf{Z}_{\leq t},\mathbf{X}_{< t},\mathbf{A}_{< t}) \big] \\
& -\mathcal{D}^\mathrm{KL}\big[ q_{\vartheta}(\mathbf{Z}_{t}|\mathbf{X}_{\leq t},\mathbf{A}_{\leq t},\mathbf{Z}_{< t}) \| p_{\theta}(\mathbf{Z}_{t}|\mathbf{X}_{< t},\mathbf{A}_{< t},\mathbf{Z}_{< t}) \big] \Big)
\end{split}
\end{align}
where $\mathbf{X}_{\leq t}$ and $\mathbf{X}_{< t}$ denote the partial sequences up to the $t^{\text{th}}$ and $({t-1})^{\text{th}}$ time samples, respectively. $\mathcal{D}^\mathrm{KL}$ denotes the (positive-valued) Kullback–Leibler divergence (KLD), which measures the statistical discrepancy of the encoder from the prior.
The conditional probabilities of (\ref{eq:sELBO_general}) encapsulate the underlying topological structure and temporal coherence of the dynamic brain networks. 
This sELBO (\ref{eq:sELBO_general}) then forms the basis of our proposed DSVB framework.

\subsection{Recurrent Graph Neural Network}

In this subsection, we introduce a model parameterization of the formulated sELBO based on a graph recurrent neural network.
Here, the conditional latent prior and approximate posterior on each node $v_i$ are modelled as Gaussian distributions:
\begin{subequations} \label{eq:Gaussian_generative}
\begin{align}
\begin{split} \label{eq:Gaussian_latent_prior}
&p_{\theta}(\mathbf{z}_{t,i}|\mathbf{X}_{< t},\mathbf{A}_{< t},\mathbf{Z}_{< t}) =  
\mathcal{N} \big( \boldsymbol{\mu}^{\text{prior}}_{t,i} , \boldsymbol{\Sigma}^{\text{prior}}_{t,i} \big) 
\end{split} \\
\begin{split} \label{eq:Gaussian_latent_posterior}
&q_{\vartheta}(\mathbf{z}_{t,i}|\mathbf{X}_{\leq t},\mathbf{A}_{\leq t},\mathbf{Z}_{< t}) = 
\mathcal{N} \big( \boldsymbol{\mu}^{\text{enc}}_{t,i} , \boldsymbol{\Sigma}^{\text{enc}}_{t,i} \big)
\end{split}
\end{align}
\end{subequations}
with isotropic covariances $\boldsymbol{\Sigma}^{\text{prior}}_{t,i} = \text{Diag}\big({\boldsymbol{\sigma}_{t,i}^{\text{prior}}}^2\big)$, $\boldsymbol{\Sigma}^{\text{enc}}_{t,i} = \text{Diag}\big({\boldsymbol{\sigma}_{t,i}^{\text{enc}}}^2\big)$, where $\text{Diag}(\cdot)$ denotes the diagonal function. The network-wide prior and approximate posterior mean and standard deviation pairs of $\mathbf{Z}_{t}$ are then obtained as follows:
\begin{subequations} \label{eq:VRNN_models}
\begin{align}
\label{eq:VRNN_prior}
\big( &\boldsymbol{\mu}^{\text{prior}}_{t} , \boldsymbol{\sigma}^{\text{prior}}_{t} \big) = \varphi_{\theta}^{\text{prior}} (\mathbf{H}_{t}) \\
\label{eq:VRNN_encoder}
\big( &\boldsymbol{\mu}^{\text{enc}}_{t} , \boldsymbol{\sigma}^{\text{enc}}_{t} \big) = \Phi_{\vartheta}^{\text{enc}} \big( \varphi_{\theta}^{x}(\mathbf{X}_{t}),\mathbf{H}_{t},\mathbf{A}_{t} \big)
\end{align}
\end{subequations}
where the prior model $\varphi_{\theta}^{\text{prior}}$, data feature model $\varphi_{\theta}^{x}$, and latent feature model $\varphi_{\theta}^{z}$ are modelled via fully-connected neural networks (FCNNs); the encoder model $\Phi_{\vartheta}^{\text{enc}}$ is modelled via the more expressive GNN.
The memory-embedding recurrent states $\mathbf{H}_{t}=[\mathbf{h}_{t,1},\ldots,\mathbf{h}_{t,N}]^{\top} \in \mathbb{R}^{N\times D_h}$ are obtained as
\begin{align} \label{eq:VRNN_rnn}
&\mathbf{H}_{t} = \Phi_{\theta}^{\mathrm{\text{rnn}}} \big(\varphi_{\theta}^{x}(\mathbf{X}_{t-1}), \varphi_{\theta}^{z}(\mathbf{Z}_{t-1}), \mathbf{H}_{t-1}, \mathbf{A}_{t-1}\big)
\end{align}
and the recurrent model $\Phi_{\theta}^{\text{rnn}}$ is modelled as a spatial-aware Gated Recurrent Unit (GRU). By virtue of (\ref{eq:VRNN_rnn}), $\mathbf{H}_{t}$ thus serves as the memory embeddings for the preceding graphs and embeddings (history path) $\mathbf{Z}_{< t}$, $\mathbf{X}_{< t}$, and $\mathbf{A}_{< t}$.
Based on our model parameterization, a deterministic closed-form solution of the KLD in sELBO (\ref{eq:sELBO_general}) can be analytically obtained as
\begin{align} \label{eq:sELBO_KLD}
\begin{split}
& \mathcal{D}^\mathrm{KL}(\theta,\vartheta) \\
&= \frac{1}{2} \sum_{i=1}^{N} \sum_{l=1}^{D_{z}} \! \left[ \frac{{{\boldsymbol{\sigma}}^{\text{enc}}_{t,il}}^{2}}{{{\boldsymbol{\sigma}}^{\text{prior}}_{t,il}}^{2}} - \log \frac{{{\boldsymbol{\sigma}}^{\text{enc}}_{t,il}}^{2}}{{{\boldsymbol{\sigma}}^{\text{prior}}_{t,il}}^{2}} + \frac{({\boldsymbol{\mu}}^{\text{enc}}_{t,il} - {\boldsymbol{\mu}}^{\text{prior}}_{t,il})^{2}}{{{\boldsymbol{\sigma}}^{\text{prior}}_{t,il}}^{2}} - 1 \right]
\end{split}
\end{align}
where $\mathbf{a}_{l}$ denotes
the $l^\text{th}$ element of the vector $\mathbf{a}$.

Subsequently, we model the adjacency matrix decoder $p_{\theta}(\mathbf{A}_{t}|\mathbf{Z}_{\leq t},\mathbf{X}_{< t},\mathbf{A}_{< t})$ as a Bernoulli distribution, conditioned on the adjacency matrix reconstruction $\mathbf{\hat{A}}_{t}$ obtained via the following inner product:
\begin{align}
\begin{split}
\label{eq:Inner_product_decoder}
\mathbf{\hat{A}}_{t} = \sigma \big( [\mathbf{Z}_{t}, \mathbf{H}_{t}] \, [\mathbf{Z}_{t}, \mathbf{H}_{t}]^{\top} \big)
\end{split}
\end{align}
where $\sigma$ denotes the sigmoid activation.
In particular, we allow the adjacency matrix decoder to be conditioned on the entire history path, which suggested the incorporation of the memory-embedding $\mathbf{H}_{t}$ in the inner-product decoder, as opposed to the variational graph recurrent neural network \cite{Hajiramezanali},
which only considered $\mathbf{Z}_{t}$. 
In consideration of (\ref{eq:Inner_product_decoder}), the adjacency matrix decoder can be written in the form of a binary cross entropy (BCE) loss as follows:
\begin{align} \label{eq:ADJ_BCE}
\begin{split}
&\mathcal{L}^{\mathrm{BCE}}(\theta, \vartheta) = \sum_{i = 1}^{N} \sum_{j = 1}^{N} \Big[ \mathbf{A}_{t,ij} \log \sigma \left( [\mathbf{Z}_{t}, \mathbf{H}_{t}]_{i}^{\top} [\mathbf{Z}_{t}, \mathbf{H}_{t}]_{j} \right) \\
&\quad + (1 - \mathbf{A}_{t,ij}) \log \left(1 - \sigma \left( [\mathbf{Z}_{t}, \mathbf{H}_{t}]_{i}^{\top} [\mathbf{Z}_{t}, \mathbf{H}_{t}]_{j} \right) \right) \Big]
\end{split}
\end{align}
where $\mathbf{A}_{ij}$ and $\mathbf{A}_{i}$
denote
the $(i,j)^\text{th}$ element and $i^\text{th}$ row of the matrix $\mathbf{A}$, respectively.
Hence, this BCE acts as the reconstruction loss between the estimated and the ground-truth adjacency graph edges.
In particular, we approximate the first expectation term in (\ref{eq:sELBO_general}) via Monte Carlo integration with Monte Carlo samples $\{\mathbf{Z}_{t}^{k},\mathbf{H}_{t}^{k}\}_{k=1}^{M}$, where $k$ is the sample index and $M$ is the number of samples.
The approximate posterior latent samples $\mathbf{Z}^{k}_{t} \sim q_{\vartheta}(\mathbf{Z}_{t}|\mathbf{X}_{\leq t},\mathbf{A}_{\leq t},\mathbf{Z}_{< t})$ are obtained via (\ref{eq:VRNN_encoder}) using the reparameterization
$\mathbf{Z}^{k}_{t} = \boldsymbol{\mu}^{\text{enc}, k}_{t} + \boldsymbol{\sigma}^{\text{enc}, k}_{t} \odot \boldsymbol{\epsilon}^{k}_{t}$, where $\boldsymbol{\epsilon}_{t}^{k} \sim \mathcal{N}(\mathbf{0}, \mathbf{I})$ and $\odot$ denotes the Hadamard (element-wise) product. The recurrent state samples $\mathbf{H}^{k}_{t}$ are obtained via (\ref{eq:VRNN_rnn}) given the previous time-step $\mathbf{Z}^{k}_{t-1}$ and $\mathbf{H}^{k}_{t-1}$.
Substituting (\ref{eq:ADJ_BCE}) and (\ref{eq:sELBO_KLD}) into (\ref{eq:sELBO_general}), we establish an unsupervised sELBO loss that underpins the proposed DSVB framework. 

In sum, the KLD (\ref{eq:sELBO_KLD}) compels the learnable encoder and prior model distributions ($p_{\theta}$, $q_{\vartheta}$) to resemble
each other, thus regularize the models from the brain network data $(\mathbf{X}_{t}, \mathbf{A}_{t})$ at current time-step. Furthermore, it promotes emphasis of the learned dynamic latent embeddings on the preceding history path $(\mathbf{X}_{< t}$, $\mathbf{A}_{< t})$ to incorporate strong temporal coherence, essential for propagating the learnable model distributions forward in time. On the other hand, the BCE (\ref{eq:ADJ_BCE}) and the inner-product decoder (\ref{eq:Inner_product_decoder}) enforce
the encoder model to construct
latent embeddings that closely adhere to the non-Euclidean spatial characteristics of the dynamic brain networks.
Together, these losses facilitate an unsupervised learning of generalized spatiotemporal embeddings that accommodate an extensive range of subject-level spatiotemporal variability across dynamic brain networks.

\subsection{Attention-based Message Passing and Spatial-aware Gated Recurrent Unit}

GNN is a class of message passing neural networks that uses aggregated topological information to construct and update node-level graph embeddings. 
In our proposed DSVB, we consider a GNN with message passing scheme inspired by the multi-head attention mechanism in the Transformer \cite{Vaswani}, of which the node embeddings are updated across layers as follows:
\begin{align} \label{eq:transconv}
\mathbf{f}_i^{(l+1)}=\mathbf{W}_1 \mathbf{f}_i^{(l)} + \sum_{j\in\mathcal{N}(i)}\alpha_{ij}\big(\mathbf{W}_2 \mathbf{f}_j^{(l)}+\mathbf{W}_5 \mathbf{f}_{ij}\big)
\end{align}
and $\alpha_{ij}$ are the attention coefficients computed as
\begin{align} \label{eq:attention}
\alpha_{ij}=\text{softmax}\left(
\frac{\big(\mathbf{W}_3 \mathbf{f}_i^{(l)}\big)^{\top} \big(\mathbf{W}_4 \mathbf{f}_j^{(l)}+\mathbf{W}_5 \mathbf{f}_{ij}\big)}{\sqrt{D_{l+1}}}
\right)
\end{align}
with learnable weights $\{\mathbf{W}_1,\mathbf{W}_2,\dots,\mathbf{W}_5\}$ that project the embeddings from $\mathbb{R}^{D_{l}}$ to $\mathbb{R}^{D_{l+1}}$, and $\mathbf{f}^{(l)}$ denotes the node embeddings at the $l^\text{th}$ GNN layer.

It follows from the scaled dot product (\ref{eq:attention}) that the node embeddings $\mathbf{f}_i$ and its neighbors $\mathbf{f}_j$ can be interpreted as the `query' and `key' of a self-attention mechanism.
The edge features are added to the key vectors in order to provide an additional source of information for the node. A similarity measure between the query and keys is then calculated via dot product to obtain the coefficients $\alpha_{ij}$, which act as attention scores in the feature aggregation step of (\ref{eq:transconv}).
The encoder model (\ref{eq:VRNN_encoder}) of the DSVB is thus modelled by a two-layered GNN with this message passing scheme, and we let $\mathbf{f}^{(0)} = \big[ \varphi_{\theta}^{x}(\mathbf{X}_{t}),\mathbf{H}_{t},\mathbf{A}_{t} \big]^{\top}$ and $\mathbf{f}^{(2)} = \big[ \boldsymbol{\mu}^{\text{enc}}_{t} ,\, \boldsymbol{\Sigma}^{\text{enc}}_{t} \big]^{\top}$ in (\ref{eq:transconv}).

To accurately model spatiotemporal dependencies, we parameterize the recurrent model (\ref{eq:VRNN_rnn}) using a spatial-aware GRU, for which the recurrent states are updated across time-steps as follows:
\begin{align} 
\label{eq:GT-GRU}
\begin{split}
&\mathbf{S}_t = \sigma \big( \Phi_{xz}(\mathbf{\bar{X}}_{t-1}, \mathbf{A}_{t-1}) + \Phi_{hz}(\mathbf{H}_{t-1}, \mathbf{A}_{t-1}) \big) \\
&\mathbf{R}_t = \sigma \big( \Phi_{xr}(\mathbf{\bar{X}}_{t-1}, \mathbf{A}_{t-1}) + \Phi_{hr}(\mathbf{H}_{t-1}, \mathbf{A}_{t-1}) \big) \\
&\tilde{\mathbf{H}}_t = \text{tanh} \big( \Phi_{xh}(\mathbf{\bar{X}}_{t-1}, \mathbf{A}_{t-1}) + \Phi_{hh}(\mathbf{R}_t \odot \mathbf{H}_{t-1}, \mathbf{A}_{t-1}) \big) \\
&\mathbf{H}_t = \mathbf{S}_t \odot \mathbf{H}_{t-1} + (1 - \mathbf{S}_t) \odot \tilde{\mathbf{H}}_t
\end{split}
\end{align}
with ${\mathbf{\bar{X}}}_{t-1} = \big[ \varphi_{\theta}^{x}(\mathbf{X}_{t-1}), \varphi_{\theta}^{z}(\mathbf{Z}_{t-1}) \big]^{\top}$.
By virtue of the model equations (\ref{eq:GT-GRU}), $\mathbf{H}_t$ thus serves as a memory embedding that retains graph-structured temporal information of the preceding latent state sequence $\mathbf{Z}_{< t}$.
In comparison to conventional GRU, the FCNN therein is replaced by the set of GNNs $\Phi_{xz}, \Phi_{hz}, \Phi_{xr}, \Phi_{hr}, \Phi_{xh}, \Phi_{hh}$. Similarly, these single-layered GNNs adopt the message passing scheme (\ref{eq:transconv}), which allows the modified GRU to simultaneously exploit meaningful spatial structures and temporal dependencies of the dynamic graph-structured data.

\subsection{Latent Embeddings for Graph Classification}
Due to the recurrent nature of our proposed DSVB, the generated latent embeddings preserve non-Euclidean topological structure and rich temporal patterns which can be further exploited for dynamic graph classification. In combination, these node embeddings entail crucial spatiotemporal information of the entire dynamic graph sequence. To summarize the subject-level dynamic graphs, we apply a flattening operation on the mean of hierarchical embeddings ${\mathbf{Z}_{T}, \mathbf{H}_{T}}$ of the final time-step $T$ to obtain the vectorized readout $\text{vec}\big( [ \boldsymbol{\mu}^{\text{enc}}_{T}, \boldsymbol{\mu}^{\text{rec}}_{T} ]^{\top} \big)$, where $\boldsymbol{\mu}^{\text{rec}}_{T} = \frac{1}{M} \sum_{k=1}^{M} \mathbf{H}^{k}_{T}$.

\begin{table*}[!th]
\begin{align} \label{eq:DSVB_final} \tag{15}
\begin{split}
&\;\mathcal{L}^\mathrm{DSVB-FCNN}(\theta,\vartheta,\tau) \;=\; \mathcal{L}^\mathrm{BCE}(\theta,\vartheta) \;+\; \mathcal{D}^\mathrm{KL}(\theta,\vartheta) \;+\; \mathcal{L}^\mathrm{MCE}(\tau) \\
&\;= \frac{1}{M} \sum_{k=1}^{M} \sum_{t=0}^{T} 
\underbrace{
\sum_{i=1}^{N} \sum_{j=1}^{N} \bigg[ \mathbf{A}_{t,ij} \log \, \sigma \big( [\mathbf{Z}^{k}_{t}, \mathbf{H}^{k}_{t}]_{i}^{\top} [\mathbf{Z}^{k}_{t}, \mathbf{H}^{k}_{t}]_{j} \big) - (1 - \mathbf{A}_{t,ij}) \log \big(1 - \sigma \big( [\mathbf{Z}^{k}_{t}, \mathbf{H}^{k}_{t}]_{i}^{\top} [\mathbf{Z}^{k}_{t}, \mathbf{H}^{k}_{t}]_{j} \big) \bigg] 
}_{\mathcal{L}^\mathrm{BCE}(\theta,\vartheta)} \\
&\quad\;+ \frac{1}{2} \sum_{t=0}^{T}
\underbrace{
\sum_{i=1}^{N} \sum_{l=1}^{D_{z}} \left[ \frac{{\boldsymbol{\sigma}^{\text{enc}}_{t,il}}^{2}}{{\boldsymbol{\sigma}^{\text{prior}}_{t,il}}^{2}} - \log \frac{{\boldsymbol{\sigma}^{\text{enc}}_{t,il}}^{2}}{{\boldsymbol{\sigma}^{\text{prior}}_{t,il}}^{2}} + \frac{(\boldsymbol{\mu}^{\text{enc}}_{t,il} - \boldsymbol{\mu}^{\text{prior}}_{t,il})^{2}}{{\boldsymbol{\sigma}^{\text{prior}}_{t,il}}^{2}} - 1 \right] 
}_{\mathcal{D}^\mathrm{KL}(\theta,\vartheta)}
\;+\; 
\underbrace{
\sum_{n=1}^{C} \mathbf{c}_{n} \log \frac{\text{exp}(\mathbf{\hat{y}}_n)}{\sum_{m=1}^{C} \text{exp}(\mathbf{\hat{y}}_m)}
}_{\mathcal{L}^\mathrm{MCE}(\tau)}
\end{split}
\end{align}
\end{table*}

Subsequently, the flattened readout is fed into the following classification model: 
\begin{align} \label{eq:classifier}
\begin{split}
\mathbf{\hat{y}} = \varphi_{\tau}^{\text{classifier}} \big( \text{vec} \begin{bmatrix} \boldsymbol{\mu}^{\text{enc}}_{T}, \boldsymbol{\mu}^{\text{rec}}_{T} \end{bmatrix}^{\top} \big)
\end{split}
\end{align}
modelled by multilayer FCNN. A softmax activation is then applied on the obtained logits $\mathbf{\hat{y}} \in \mathbb{R}^2$ to get the predictive probability scores for the final subject-level brain network classification. The predicted class label is thus the one with highest predictive score.
The classification loss is taken to be the following multi-class cross entropy:
\begin{align} \label{eq:CL_CE}
\begin{split}
\mathcal{L}^\mathrm{MCE}(\tau) = \sum_{n=1}^{C} \mathbf{c}_{n} \log \frac{\text{exp}({\hat{y}}_n)}{\sum_{m=1}^{C} \text{exp}({\hat{y}}_m)}
\end{split}
\end{align}
where $\mathbf{c}\in \{0,1\}^C$ is the class label and $C$ is the number of classes.
Incorporating (\ref{eq:CL_CE}) into the sELBO (\ref{eq:sELBO_general}), we obtain the final DSVB-FCNN loss $\mathcal{L}^\mathrm{DSVB}$ in (\ref{eq:DSVB_final}). 

Unlike previous methods, the FCNN classifier (\ref{eq:classifier}) is trained jointly with the variational recurrent graph autoencoder (\ref{eq:VRNN_models})-(\ref{eq:VRNN_rnn}) in an end-to-end fashion. Gradients of the classifier are allowed to back-propagate through time and update parameters $\{\theta,\vartheta\}$ of the DSVB models.

\subsection{Adversarial Model Regularization}

Under a limited amount of training data, deep graph representation learning models are prone to generating node embeddings that aggravate overfitting. Such models typically construct a latent embedding space that overly- or grossly - overfits the limited dataset. Consequently, downstream task-specific (e.g., graph classification) models that are adapted to this data-specific latent space perform poorly on unseen latent embeddings of general graphs. Taking this into consideration, a probabilistic adversarial training strategy is introduced to regularize the latent embedding space constructed by the DSVB models. With the adversarial regularization, the DSVB models are expected to overcome the data overfitting issue by realizing a more inclusive latent embedding space that can be readily extrapolated to unseen data beyond the limited training dataset. 

Inspired by the model perturbation strategy in domain adversarial training \cite{Ganin}, we train the parameters $\{\theta,\vartheta,\tau\}$ of the DSVB-FCNN loss (\ref{eq:DSVB_final}) in adversarial fashion:
\begin{align} \label{eq:adverserial_objective} \tag{16}
\begin{split}
(\theta,\vartheta) &= \argmax_{\theta,\vartheta} \; \mathcal{L}^\mathrm{DSVB-FCNN}(\theta,\vartheta,\tau) \\
\tau &= \argmin_{\tau} \; \mathcal{L}^\mathrm{MCE}(\tau)
\end{split}
\end{align}
On one hand, the DSVB parameters $(\theta,\vartheta)$ are optimized to generate hierarchical embeddings $\mathbf{Z}_{T}, \mathbf{H}_{T}$ that fool the classifier $\varphi_{\tau}^{\text{classifier}}$.
On the other hand, the FCNN classification model parameters $\tau$ are optimized to accurately distinguish the class label of the generated latent embeddings.
Such an adversarial competition is expected to achieve the Nash equilibrium that generalizes the latent embedding space and prevents model overfitting.

\begin{table*}[tb]
\centering
\resizebox{1\textwidth}{!}{
\begin{tabular}{lllllll}
\hline \hline
Type of FC  &  Classifier & Accuracy (\%)  & Recall (\%) & Precision (\%) & F1-Score (\%) & AUC (\%)\\ \hline
\multirow{5}{*}{Static}  
& SVM & 56.26 $\pm$ 4.51 & 55.71 $\pm$ 3.19 & 55.05 $\pm$ 4.69 & 55.37 $\pm$ 3.97 & 56.24 $\pm$ 3.99\\
& BrainNetCNN \cite{kawahara2017brainnetcnn}   & 57.04 $\pm$ 10.99 & 44.29 $\pm$ 15.29 & 61.52 $\pm$ 15.24 & 49.03 $\pm$ 13.10 & 56.62 $\pm$ 10.86\\
& ASD-DiagNet \cite{eslami2019asd}	& 68.03 $\pm$ 2.87 & 58.57 $\pm$ 12.29	& 74.04 $\pm$ 10.60 & 63.55 $\pm$ 4.05 & 67.62 $\pm$ 2.64\\
& GroupINN \cite{yan2019groupinn}	& 64.53 $\pm$ 9.06 & 34.29 $\pm$ 24.49	& 67.88 $\pm$ 35.09	& 43.10 $\pm$ 25.83 & 63.76 $\pm$ 9.37\\
& Hi-GCN \cite{jiang2020hi}	& 66.60 $\pm$ 6.05	& 60.00 $\pm$ 10.69	& 68.51 $\pm$ 7.19	& 63.30 $\pm$ 6.74 & 66.38 $\pm$ 6.06\\
& E-Hi-GCN \cite{li2021te}	& 66.60 $\pm$ 8.64	& 54.29 $\pm$ 21.48	& 75.42 $\pm$ 14.71 & 58.31 $\pm$ 18.40 & 58.00 $\pm$ 6.38\\ \hline
\multirow{4}{*}{Dynamic}
& SVM & 63.82 $\pm$ 9.41& 55.43 $\pm$ 7.23& 56.66 $\pm$ 3.34 & 55.52 $\pm$ 3.61 & 63.52 $\pm$ 8.20\\
& BrainNetCNN \cite{kawahara2017brainnetcnn} & 54.49 $\pm$ 8.15 & 57.14 $\pm$ 8.34 & 50.00 $\pm$ 14.00 & 53.33 $\pm$ 10.15 & 54.58 $\pm$ 7.03\\
& GAE-FCNN \cite{fuad2022} & 66.03 $\pm$ 7.14 & 65.71 $\pm$ 14.57 & 68.92 $\pm$ 16.26 & 64.96 $\pm$ 6.37 & 66.19 $\pm$ 7.17\\
& GAE-LSTM \cite{fuad2022} & 54.78 $\pm$ 6.05 & 47.14 $\pm$ 19.48 & 71.00 $\pm$ 25.12 & 54.92 $\pm$ 17.09 & 54.53 $\pm$ 6.13\\
& \textbf{DSVB-FCNN} & \textbf{78.44 $\pm$ 2.77} & \textbf{66.67 $\pm$ 7.45} & \textbf{89.64 $\pm$ 5.90} & \textbf{75.99 $\pm$ 4.01} & \textbf{78.94 $\pm$ 2.64}\\
\hline \hline
\end{tabular}}
\caption{Performance comparison of our proposed DSVB-FCNN with different static and dynamic baseline classifiers.}
\label{table:benchmark}
\end{table*}

\begin{table*}[tb]
\centering
\resizebox{\textwidth}{!}{
\begin{tabular}{rllllll}
\hline \hline
& Model Specification
& Accuracy (\%)  & Recall (\%) & Precision (\%) & F1-Score (\%) & AUC (\%)\\ \hline
I &  Graph Autoencoder + Conventional GRU & 72.39 $\pm$ 9.01  & 65.00 $\pm$ 8.16 & 78.91 $\pm$ 13.04 & 71.02 $\pm$ 9.05 & 72.65 $\pm$ 9.08\\
II & Graph Autoencoder + Spatial-aware GRU & 76.63 $\pm$ 9.08 & 73.33 $\pm$ 8.16 & 82.05 $\pm$ 11.47 & 76.74 $\pm$ 7.47 & 76.74 $\pm$ 9.32\\
III & Variational Graph Autoencoder + Spatial-aware GRU & 75.00 $\pm$ 10.06 & 63.33 $\pm$ 6.67 & 86.94 $\pm$ 15.26 & 72.78 $\pm$ 8.90 & 75.38 $\pm$ 10.31\\
IV & Graph Autoencoder + Spatial-aware GRU + Adverserial Training & 77.61 $\pm$ 5.65 & \textbf{76.67 $\pm$ 6.24} & 79.48 $\pm$ 6.45 & \textbf{77.96 $\pm$ 5.73} & 77.65 $\pm$ 5.67\\
V & Variational Graph Autoencoder + Spatial-aware GRU + Adverserial Training & \textbf{78.44 $\pm$ 2.77} & 66.67 $\pm$ 7.45 & \textbf{89.64 $\pm$ 5.90} & 75.99 $\pm$ 4.01 & \textbf{78.94 $\pm$ 2.64}\\ \hline \hline
\end{tabular}}
\caption{Ablation study of our proposed DSVB framework based on a nested-stratified 5-fold cross-validation.}
\label{table:ablation}
\end{table*}

\section{Application to Dynamic Brain Networks}

\subsection{fMRI Dataset \& Preprocessing}

We used a C-PAC pipeline \cite{Cameron2013-nn} preprocessed fMRI dataset from the Autism Brain Imaging Data Exchange (ABIDE I) open source database \cite{di2009functional}, with a sample of 144 subjects (70 ASD and 74 healthy controls (HC)) resting-state fMRI included in this case study. The inclusion criteria \cite{Plitt2015} are: males with a full-scale IQ $>$ 80; ages between 11 and 23; and the fMRI acquisition sites are New York University,
University of California Los Angeles 1, and University of Utah, School of Medicine. We performed parcellation using Power et al.'s brain atlas \cite{Power2011-hs} to extract the mean time series for a total of $N=264$ regions of interest (ROIs). 

\subsection{Graph Neural Network Model Training}

The proposed DSVB framework is used to generate the dynamic node embeddings that correspond to the node feature and adjacency matrix sequences. The FCNN classifier is then used to discriminate between the ASD and HC subjects. We train the end-to-end DSVB-FCNN via an Adam optimizer using a learning rate of 1e-5 = $10^{-5}$ with exponential annealing, and a L2 regularization of 0.01 for 400 epochs. The output dimension of the graph-structure GRU is chosen to be 16. The output dimensions of each layer of the GNN encoder are chosen to be 32 and 16, respectively. 
The output dimensions of the single-layered FCNN data and latent feature models are chosen to be 64 and 8, respectively. The output dimensions of each layer of the FCNN classifier are chosen to be 32 and 2, respectively. These model hyperparameters are selected based on a 5-fold cross-validation.

\subsection{Baseline Methods}

\noindent \textbf{SVM}: The flattened upper triangle of the FC matrix is used as input to SVM.
\noindent \textbf{BrainNetCNN} \cite{kawahara2017brainnetcnn}: An extension of CNNs to handle brain graph-structured data using special kernels.  
\noindent \textbf{GroupINN} \cite{yan2019groupinn}: An ensemble of GCNs to learn graph-level latent embedding representations.
\noindent \textbf{ASD-DiagNet} \cite{eslami2019asd}: Dense autoencoder for embedding learning and single-layered perceptron for classification.
\noindent \textbf{Hi-GCN} \cite{jiang2020hi}: This method uses GroupINN to produce the embedding for all network instances, the learned embeddings are fed to a population-based GCN for classification.
\noindent \textbf{E-HI-GCN} \cite{li2021te}: An ensemble of HI-GCN, each of which is trained on different sparsity level brain networks. 
\noindent \textbf{GAE-FCNN} \cite{fuad2022}: GAE is used to learn dynamic graph-level latent embeddings, which are then fed to a FCNN for classification. 
\noindent \textbf{GAE-LSTM} \cite{fuad2022}: The learned embeddings of GAE are fed to a LSTM for classification.

\section{Results and Discussion}
To evaluate the effectiveness of our proposed method, a nested-stratified 5-fold cross-validation was applied for each experiment.
The performance of ASD identification is measured using five metrics: accuracy, recall, precision, F1-score, and area under the curve (AUC).
Table \ref{table:benchmark} compares the average performances of the proposed DSVB-FCNN with the existing baseline classifiers based on both dynamic and static FC networks. It shows that our proposed DSVB-FCNN achieves substantial improvement over the baseline classifiers in each metric.
The DSVB-FCNN also achieved low standard deviations in all metrics, demonstrating a strong consistency in its performances across the 5-fold cross-validation. In general, these results indicate that the framework facilitates robust brain disorder identification by generating graph latent embeddings that generalize well to unseen dynamic FC networks beyond the limited training dataset.

To provide insights into these performance gains of the DSVB-FCNN, we conduct an ablation study where certain components (i.e., variational Bayes, spatial-aware GRU and adversarial training) of the DSVB framework are removed to highlight their distinctive contributions to the improved dynamic FC classification result. Table \ref{table:ablation} shows the average performances of the ablation study based on five DSVB model variants. 
Model I underperformed in comparison to other spatial-aware GRU-empowered models. This indicates the importance of the proposed spatial-aware GRU in endowing recurrent model's ability to capture rich spatiotemporal patterns crucial for dynamic network classification.

\begin{figure}[t]
\centerline{\includegraphics[width=0.725\columnwidth]{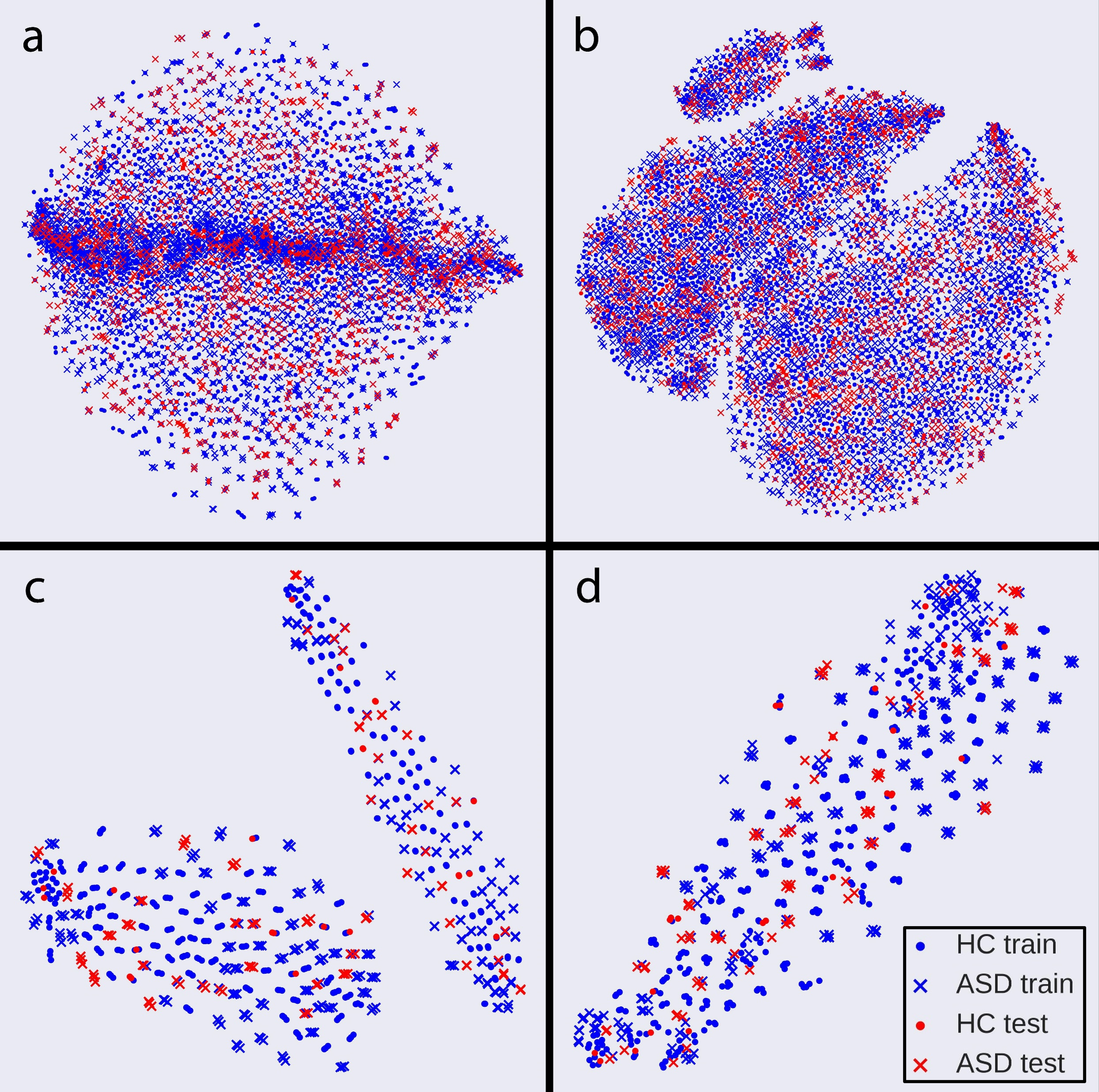}}
\caption{Top figures are t-SNE visualizations on the latent state sequences $\{\mathbf{Z}_{t}\}_{t=0}^{T}$ of (a) graph recurrent autoencoder and (b) variational graph recurrent autoencoder. Bottom figures are t-SNE visualizations on the final (time-step) readouts $\text{vec}([\boldsymbol{\mu}^{\text{enc}}_{T}, \boldsymbol{\mu}^{\text{H}}_{T}]^{\top})$ of (c) variational graph recurrent autoencoder and (d) variational graph recurrent autoencoder with adversarial training.}
\label{fig:tsne}
\end{figure}

\begin{figure*}[th]
\centerline{\includegraphics[width=0.9\textwidth]{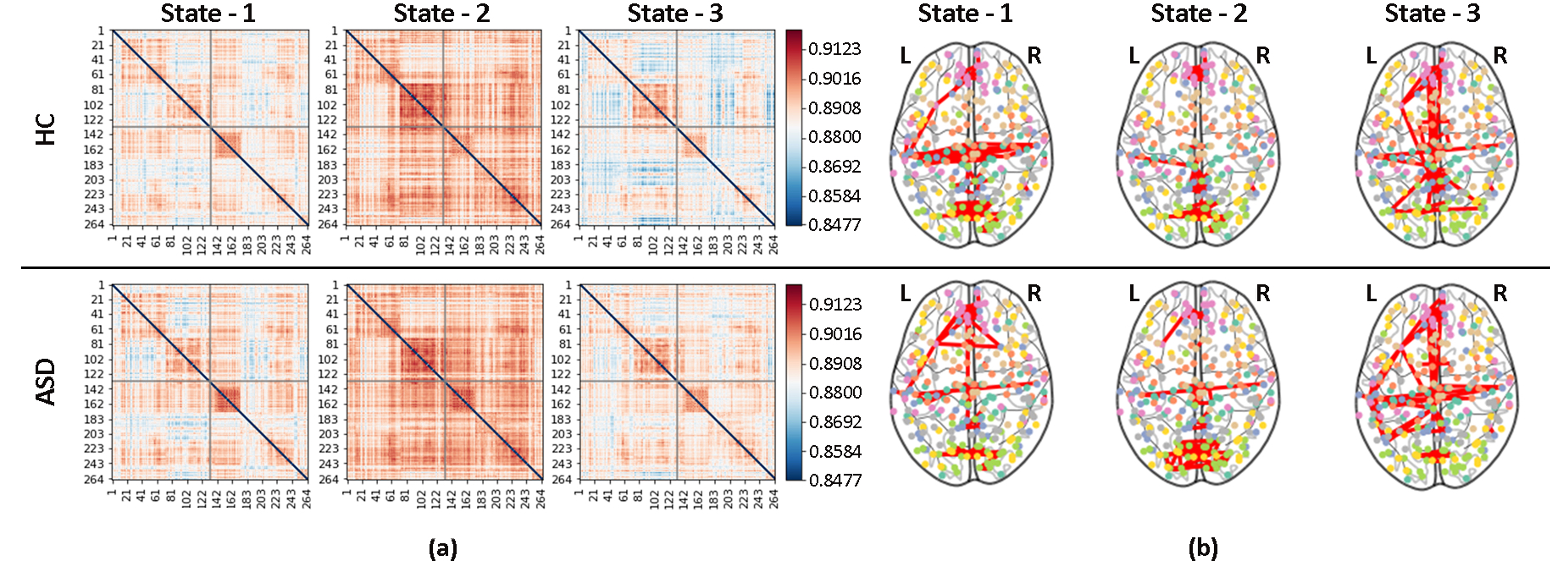}}
\caption{The average connectivity patterns derived from DSVB-learned embeddings. (a) 3 main states derived from k-means clustering of 264 x 264 ROIs connectivity matrices from HC and ASD groups. (b) The corresponding connectome plots for HC and ASD.}
\label{fig:connectivity}
\end{figure*}

Figures \ref{fig:tsne}a and \ref{fig:tsne}b show the t-distributed Stochastic Neighbor Embedding (t-SNE) visualizations of respective latent state sequences $\{\mathbf{Z}_{t}\}_{t=0}^{T}$ from models II and III of the ablation study. It shows that the latent state t-SNEs of model II are more concentrated towards the centre in comparison to the t-SNEs of model III,
which is more uniform.
This indicates that the VB elements (KLD and reparameterization) facilitate learning of a smooth latent embedding space to better accommodate subject-level spatiotemporal variability across unseen dynamic FC networks. 
Nevertheless, due to incorporation of the prior model via KLD loss (\ref{eq:sELBO_KLD}), VB-based models generate latent embeddings that are highly accustomed to the spatiotemporal structure of the dynamic FC data. This aggravates model overfitting and increases the chance of false classification, resulting in an inferior performance of model III in Table \ref{table:ablation} compared to model II.

Figures \ref{fig:tsne}c and \ref{fig:tsne}d show the t-SNE visualizations of respective readouts $\text{vec}([\boldsymbol{\mu}^{\text{enc}}_{T}, \boldsymbol{\mu}^{\text{rec}}_{T}]^{\top})$ from models III and V.
In particular, Figure \ref{fig:tsne}c shows that model III divided the readout t-SNEs into two predominant clusters based on the causal structure of each dynamic FC data.
In contrast, t-SNEs of model V in Figure \ref{fig:tsne}d are more regularized,
which demonstrates the effectiveness of the proposed adversarial training strategy in learning indiscriminate latent embedding space that generalized well to unseen dynamic FC networks. The coordination between all components of the DSVB framework produces superior ASD identification performance,
as shown in Tables \ref{table:benchmark} and \ref{table:ablation}.

\begin{figure}[H]
\centerline{\includegraphics[width=0.75\columnwidth]
{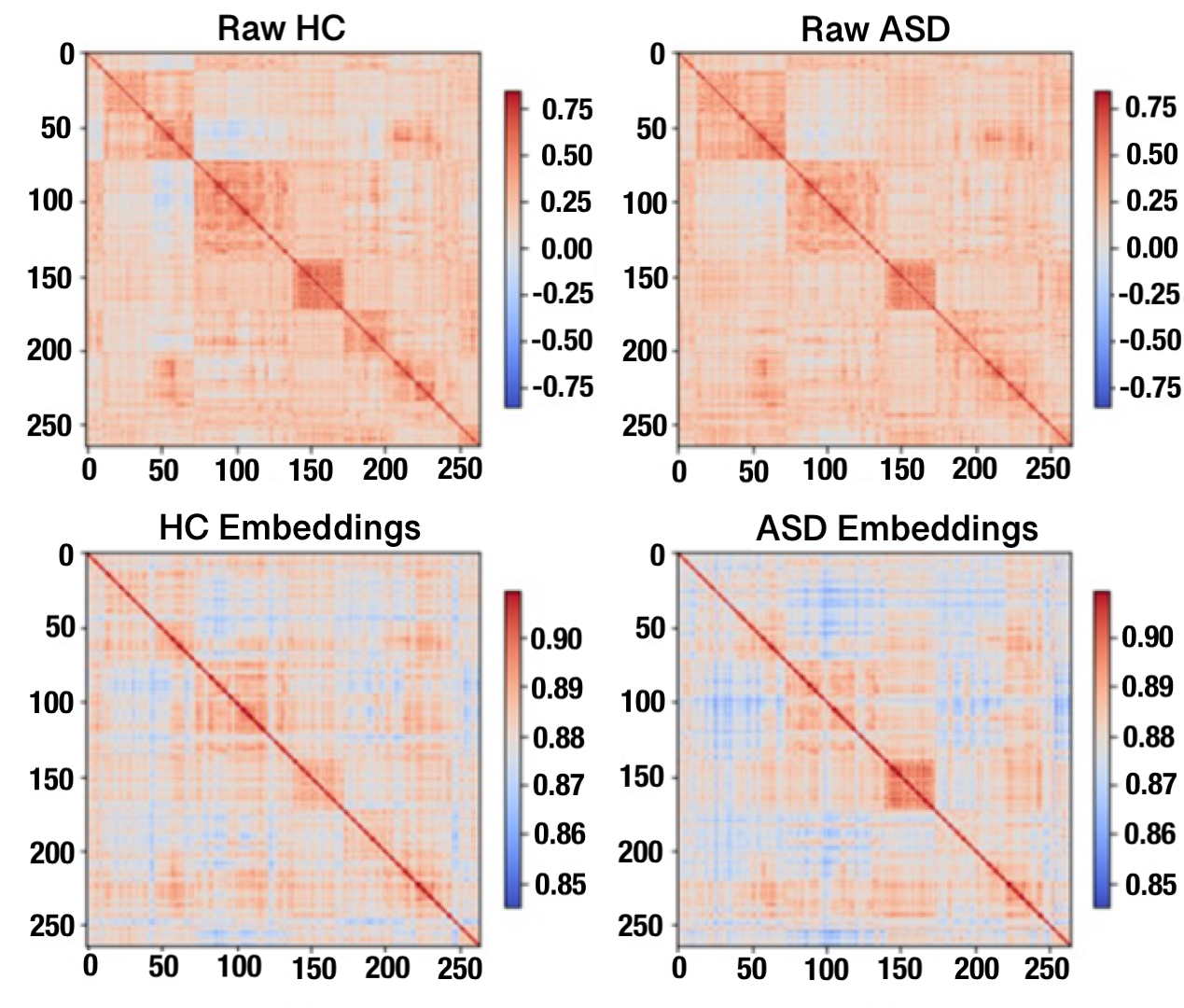}}
\caption{Comparison of the group means of raw FC (264 x 264 ROIs) with the latent embeddings derived from our DSVB framework over time for both HC and ASD groups.}
\label{fig:raw_vs_embeddings}
\end{figure}

In addition, we constructed higher-order dynamic FC by correlating the DSVB-learned embeddings $\mathbf{z}_{it}$ between pairs of nodes, and examined the dynamic connectivity states in ASD by applying dynamic FC clustering of these dynamic networks. The optimal number of clusters = 3 is determined using the Elbow method \cite{bholowalia2014ebk}. Figure ~\ref{fig:connectivity} shows the estimated connectivity matrices for three different states with the corresponding highest FC connections. By referring to \cite{Power2011-hs}, the corresponding brain functional system can be traced back. The ASD group shows stronger FC connections in visual, default mode, salience networks; whereas higher connectivity is observed in HC for sensory and auditory networks. The difference in FC patterns between ASD and HC is state-dependent. In ASD, visual ROIs show increased interconnections in state-1 and state-2 where default mode and salience are mostly activated in state-1. This is consistent with \cite{Holiga2019-rf}, who also concluded that - compared to HC - ASD has hyperconnectivity in prefrontal and parietal cortices but hypoconnectivity in sensory-motor regions.
Figure \ref{fig:raw_vs_embeddings} shows that the latent embedding matrices derived from DSVB are more visually discriminative compared to the raw FC matrices. The ASD embeddings highlight hyperconnectivity within visual networks (ROIs 143 to 173), particularly on the occipital lobe, and this aligns with earlier neuroscience discoveries \cite{keehn2008functional,clery2013fmri,matsuoka2020increased}.

\section{Conclusion}

We have developed a deep probabilistic spatiotemporal framework based on sequential variational Bayes for dynamic graph representation learning. The proposed DSVB framework incorporates a spatial-aware GRU to capture both topological and temporal alterations across brain networks. A downstream FCNN then leverages the learned graph embeddings to reveal atypical neural connectivity patterns. Evaluation on resting-state fMRI data shows substantial improvements over the existing baseline methods, suggesting potential in neuropsychiatric disorder identification. Despite our current focus on brain disorders, this framework is domain-agnostic and applicable to other dynamic graph-structured data. Moreover, the proposed framework could be extended to handle large graphs by incorporating sparse attention and heterogeneous graphs via cross-modality representations. 

\section*{Acknowledgements}

The work of Junn Yong Loo is supported by Monash University under the SIT Collaborative Research Seed Grants 2024 I-M010-SED-000242. The work of Chee-Ming Ting is supported by the Ministry of Higher Education, Malaysia under the Fundamental Research Grant Scheme FRGS/1/2023/ICT02/MUSM/02/1. Sin-Yee Yap has been based at Monash’s Department of Data Science and AI for some of the duration of this work.

\section*{Contribution Statement}

Sin-Yee Yap, Junn Yong Loo (corresponding author) and Chee-Ming Ting contributed equally to this work.

\bibliographystyle{named}
\bibliography{ijcai24}

\end{document}